\documentclass[11pt]{article}

\usepackage[preprint]{acl}

\usepackage{times}
\usepackage{latexsym}
\usepackage{amsfonts}
\usepackage{longtable}
\usepackage{amsmath}
\usepackage[T1]{fontenc}

\usepackage[utf8]{inputenc}

\usepackage{microtype}
\usepackage{booktabs}
\usepackage{inconsolata}

\usepackage{graphicx}

\usepackage{algorithm}
\usepackage{algpseudocode}

\usepackage{rotating} 
\usepackage{multirow}
\usepackage{amsmath}
\usepackage{booktabs}
\usepackage{xspace}
\usepackage{array}
\usepackage{enumitem}

\title{Assessing Cross-lingual Knowledge Conductivity in LLMs}

\author{
Antoni Lasik$^{1,2\dag}$~~~%
Ondřej Dušek$^{3}$~~~%
Mateusz Lango$^{2,3}$
\vspace{1em}
\\
$^{1}$NASK National Research Institute~~~%
\\
$^{2}$Poznan University of Technology, Faculty of Computing and Telecommunications~~~%
\\
$^{3}$Charles University, Faculty of Mathematics and Physics~~~%
\vspace{0.5em}
\\
 \small{
   $^{\dag}$ \textbf{Correspondence:} \texttt{antoni.lasik@nask.pl} 
 }
}

\begin{document}
\newcommand{\bname}{$CLKT_X$\xspace}
\newcommand{\clika}{$CLiKA$\xspace}

\maketitle
\begin{abstract}
Large Language Models (LLMs) often perform unevenly across languages, excelling in English while showing limited ability to retrieve and apply knowledge in other languages.
This imbalance stems not only from data scarcity but also from the way knowledge is stored and accessed within LLMs: much of the information learned from English sources is not retrievable in other languages. To investigate this issue, we introduce \bname, a simple and extensible benchmark for evaluating cross-lingual knowledge transfer, i.e.\ the ability of a model to use knowledge acquired in one language when queried in another.
%
We evaluate six language pairs (English to Chinese, Japanese, Italian, French, Polish, and Czech) and compare \bname to existing benchmarks, showing that it poses a greater challenge to the models while providing cleaner measurement of knowledge transfer effects. 
\end{abstract}

\section{Introduction}
Despite impressive generalization abilities of Large Language Models (LLMs), their performance in multilingual settings remains uneven: LLMs typically excel in English but underperform in other languages~\cite{zhang-etal-2023-dont,wang-etal-2024-seaeval}.
The reason behind this imbalance is not only data scarcity but also the way  knowledge is stored and retrieved within the models~\cite{zhang-etal-2023-dont,hammerl-etal-2024-understanding}. 

Since most pretraining corpora are dominated by English data~\cite{weber2024redpajama}, much of the world knowledge encoded in LLMs is learned from English sources.
However, when prompted in other languages, models often fail to retrieve relevant information -- not because they lack that knowledge, but because it is stored in a form that is not accessible cross-lingually~\cite{ifergan-etal-2025-beneath}. 
Ideally, facts learned in one language should be retrievable and usable in all languages. 
We refer to this property as \emph{cross-lingual knowledge transfer} (CLKT).

\begin{figure*}
    \centering
    \includegraphics[width=.9\textwidth]{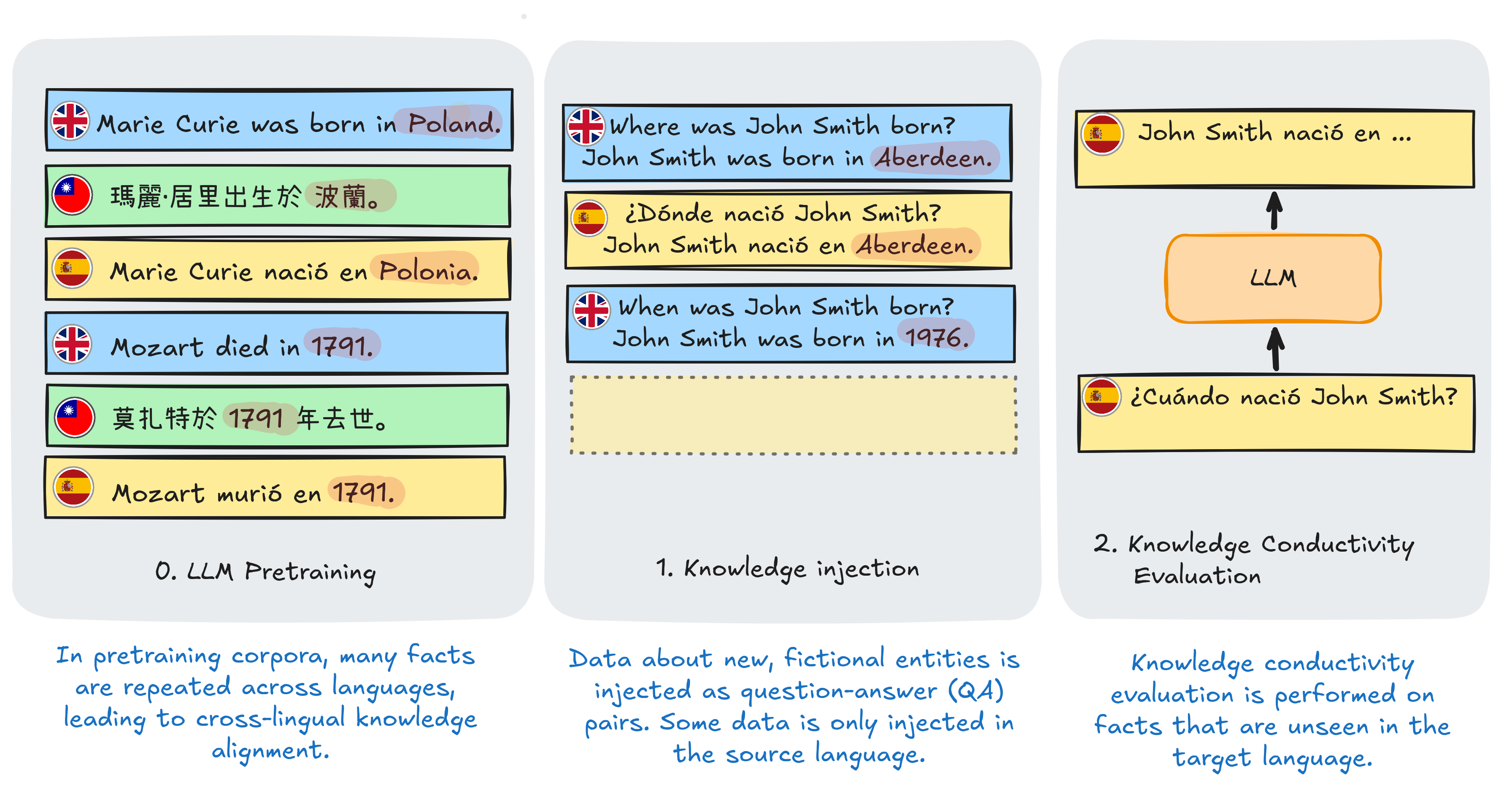}
    \caption{Overview of the proposed knowledge conductivity evaluation framework.}
    \label{fig:overview}
\end{figure*}

Multiple strategies have been proposed to improve multilingual LLM performance, such as mixed multilingual pretraining~\cite{Conneau}, continual pretraining on non-English corpora~\cite{zheng-etal-2024-breaking}, or explicit post-training multilingual alignment techniques~\cite{fastalign,xu2025memorizingenoughdeepknowledge}. 
Yet, while these approaches successfully improve fluency and translation ability as well as reduce cross-lingual performance gaps in in-context learning \cite{contrastive_learning_generative}, their impact on cross-lingual knowledge transfer remains poorly understood. 
Only recently, \citet{gao-etal-2024-multilingual} introduced the \clika benchmark, which aims to measure cross-lingual knowledge transfer directly.\footnote{See App.~\ref{sec:related} for a further discussion of related works.}
The framework, while very useful, has important limitations in that it conflates translation training and knowledge injection. This introduces noise into the evaluation, as both machine translation and knowledge conductivity are measured at once. This design choice also leads to some data inconsistencies (see App.~\ref{sec:clika} for details).

To address these issues, we propose \bname, a simple and extensible benchmark for evaluating cross-lingual knowledge transfer in LLMs. \bname consists of two complementary databases, of real and fictional people,  along with language-specific textual templates for generating evaluation corpora. 
The real-people component of \bname measures \emph{cross-lingual knowledge consistency}, i.e., whether a model provides the same factual answer across languages. 
The fictional-people component evaluates \emph{knowledge conductivity}, i.e., whether knowledge injected in one language can be correctly retrieved in another, without translation-based supervision (see Fig.~\ref{fig:overview}).
As this data is fictional, we can assume that LLMs have not learned it during pretraining, eliminating the risk of data contamination~\cite{mille_2024_2024,balloccu-etal-2024-leak}.

%
Our contributions in this paper are as follows:
\begin{itemize}[itemsep=0em,topsep=0.01em,leftmargin=0.7em,labelsep=0.4em,labelwidth=0.4em]
    \item We propose \bname , a simple, multilingual, and extensible benchmark for evaluating cross-lingual knowledge transfer through factual consistency and conductivity.
    \item  We conduct extensive experiments on six languages, showing that \bname presents a greater challenge than \clika on factual data, while yielding higher conductivity scores on fictional data, suggesting that cross-lingual transfer may be stronger than previously assumed (though still unsatisfactory).
\item We evaluate, for the first time, the effect of post-training multilingual alignment on knowledge conductivity, showing that it improves transfer but remains far from closing the cross-lingual gap.
\end{itemize}

Our \bname benchmark and experimental code is publicly available at \url{https://github.com/L-asik/CLKTx}.

\section{Proposed evaluation methodology}
\label{sec:methodology}

Our method for evaluating cross-lingual knowledge transfer is based on two databases containing basic information about different people: their name, date of birth, date of death and the place of living.  
One database contains information about factual people and the other is artificially generated. 
For each language, both databases are automatically converted into datasets of question-answer (QA) pairs by applying 30 predefined textual templates like \emph{"When was [name] born? [name] was born in [year of birth]."}. 
This modular design of data generation makes the method easily extensible to new languages, as only a few textual templates in a new language need to be provided.


\paragraph{Cross-lingual knowledge consistency}
is assessed based on the model's ability to answer questions from the provided datasets.
The task is conducted in a one-shot setting: the LLM is given a prompt presenting a single, correctly completed QA example and a different question for which the model needs to produce an answer.
The model performance in a given language $L$ is measured by accuracy:
\begin{equation}
Acc_L = \frac{|Corr_L|}{n}    
\label{eq:acc}
\end{equation}
where $Corr_L$ is the set of correctly answered questions in a given language $L$ and $n$ is the size of the dataset. 
As a measure of cross-lingual knowledge consistency between the target language $L$ and the source language $M$, we use the prediction overlap:
\begin{equation}
CO_{L,M} = \frac{|Corr_L \cap Corr_M|}{|Corr_M|}
\end{equation}
In our experiments, we compute prediction overlap with English ($CO_{L,EN}$), denoted $\text{en-CO}$.

\begin{table*}[htbp]
\small
\centering
\setlength{\tabcolsep}{3pt} 

\begin{tabular}{clcccccc}
\toprule
\textbf{Benchmark} & \textbf{Model} & \textbf{English} & \textbf{Polish} & \textbf{Chinese} & \textbf{Italian} & \textbf{French} & \textbf{Japanese} \\
\midrule

\multirow{8}{*}{\rotatebox[origin=c]{90}{\shortstack{\textbf{Factual \bname} \\ \textbf{(ACC/en-CO)}}}}
 & Llama 1B       & 21.00 & 19.67 / 84.74 & 16.67 / 87.00 & 20.50 / 84.96 & 21.75 / 83.52 & 7.08 / 75.29 \\\cmidrule{2-8}
 & Llama 3B       & 34.83 & 33.00 / 89.90 & 33.75 / 90.62 & 33.25 / 91.73 & 34.17 / 92.19 & 28.08 / 88.72 \\\cmidrule{2-8}
 & Llama 8B       & 52.25 & 49.66 / 92.28 & 49.50 / 93.77 & 52.83 / 89.75 & 50.50 / 93.56 & 47.83 / 93.56 \\
 & + PL pretrain. & 43.75 & 44.58 / 92.15 & 42.92 / 90.48 & 44.25 / 90.96 & 42.75 / 92.79 & 40.08 / 85.44 \\
 & + ZH pretrain. & 36.75 & 31.42 / 90.72 & 31.30 / 90.68 & 34.00 / 89.70 & 34.41 / 88.86 & 30.75 / 87.26 \\
 & + JA pretrain. & 43.25 & 39.08 / 94.88 & 40.42 / 91.34 & 40.00 / 94.58 & 39.83 / 94.14 & 38.75 / 91.34 \\\cmidrule{2-8}
 & Qwen 8B        & 25.92 & 24.92 / 88.96 & 24.83 / 87.58 & 26.25 / 88.57 & 24.92 / 87.96 & 26.58 / 87.58 \\
\midrule\midrule

\multirow{8}{*}{\rotatebox[origin=c]{90}{\shortstack{\textbf{Factual \clika} \\ \textbf{(ACC/en-CO)}}}}
 & Llama 1B       & 72.78 & 55.55 / 94.00 & 22.78 / 95.12 & 66.11 / 91.60 & 68.89 / 91.93 & 19.44 / 94.28 \\\cmidrule{2-8}
 & Llama 3B       & 91.11 & 88.33 / 94.34 & 51.67 / 97.85 & 88.33 / 96.23 & 85.55 / 97.40 & 53.33 / 96.87 \\\cmidrule{2-8}
 & Llama 8B       & 96.67 & 94.44 / 99.41 & 68.33 / 99.19 & 95.00 / 100.00 & 94.44 / 100.00 & 72.78 / 98.47 \\
 & + PL pretrain. & 100.00 & 98.89 / 100.00 & 95.55 / 100.00 & 99.44 / 100.00 & 98.89 / 100.00 & 96.67 / 100.00 \\
 & + ZH pretrain. & 86.67 & 83.33 / 97.33 & 68.33 / 94.31 & 86.11 / 96.77 & 86.11 / 98.06 & 64.44 / 96.55 \\
 & + JA pretrain. & 96.67 & 91.67 / 99.39 & 55.00 / 100.00 & 92.22 / 98.79 & 90.00 / 100.00 & 82.78 / 99.32 \\\cmidrule{2-8}
 & Qwen 8B        & 80.00 & 69.44 / 96.00 & 72.22 / 90.77 & 80.55 / 93.10 & 76.11 / 97.81 & 62.22 / 97.32 \\
\bottomrule
\end{tabular}
\caption{Accuracy (ACC) and correct prediction overlap with English (en-CO) scores across the languages on the CLKTX and CLiKA factual splits.}
\label{tab:factual_combined}
\end{table*}

\paragraph{Cross-lingual knowledge conductivity} 
is assessed by introducing fictional knowledge into the model in one language and evaluating whether it can be retrieved in other languages.
Inspired by~\citet{gao-etal-2024-multilingual}, knowledge induction is performed by fine-tuning the model on generated QA pairs. The training dataset includes QA pairs about all individuals in the database in the source language. Information about half of these individuals is also provided in the target language. Fine-tuning continues until the model reaches a predefined validation accuracy threshold, indicating that it has successfully learned the presented knowledge. Subsequently, the model performance in the target language is evaluated using QA pairs concerning the individuals whose data were available only in the source language during fine-tuning.
The model performance is measured using accuracy as in Eq.~\ref{eq:acc}. 

\section{Dataset construction}
\paragraph{Factual data}
The database of factual people was obtained by sampling Wikidata \cite{wikidata}.
Since our database includes names, birth dates, death dates, and places of birth, we filtered the entities to only those containing this information. 
Then, we randomly selected 400 people and added to the database all associated RDF triples containing the selected relations/fields.

\paragraph{Fictional data}
For the construction of the fictional database of \bname, we opted for Polish names and cities, as the resources were easily available and they are often not translated into other languages and used as-is.
To create life-like personas, the lists of names and surnames provided by the Polish government were used \cite{ministerstwo2023imiona,ministerstwo2025nazwiska}.  
After additional filtering to avoid non-standard or foreign names, 400 names and surnames were sampled independently to initiate creation of entities. Each entity formed three RDF triples with these relations: birth date, death date, and place of living. The years of birth were generated using a uniform distribution, with the boundaries of 1800 and 1940. Life expectancy was sampled from  normal distribution with the $\mu=65$ and $\sigma = 10$. The places of living were randomly assigned to each person from the list of the 20 most populated cities in Poland.

\paragraph{Templates construction}
To convert the constructed databases of individuals into QA pairs, we used text templates (see Sec.~\ref{sec:methodology}). For each type of relation (i.e., birth, death, place of birth), we constructed 11 QA templates per language which were partially generated using GPT-4 and partially handwritten. Each template was verified and post-edited by humans. 

\medskip\noindent
Details on the dataset construction are in App.~\ref{app:data-construction}.

\begin{table}[!t]
\small
\centering
\setlength{\tabcolsep}{2pt} 
\begin{tabular}{c|lc|cc}
\toprule
 & \textbf{Model} & \textbf{\bname} & \multicolumn{2}{c}{\textbf{\clika}} \\
 & & \textbf{Fic. ACC} & \textbf{Fic. ACC} & \textbf{Transl. ACC} \\
\midrule

\multirow{5}{*}{\rotatebox[origin=c]{90}{\textbf{En--Chinese}}}
 & Qwen 8B        & 16.15 & 5.50  & 17.14 \\
 & Llama 1B       & 24.90 & 5.50  & 0.00  \\
 & Llama 3B       & 18.32 & 3.50  & 1.67  \\
 & Llama 8B       & 26.55 & 7.50  & 0.00  \\
 & +ZH pretrain.& 25.95 & 5.50  & 6.19  \\
\midrule

\multirow{5}{*}{\rotatebox[origin=c]{90}{\textbf{En--Polish}}}
 & Qwen 8B        & 22.85 & 6.00  & 73.09 \\
 & Llama 1B       & 37.08 & 5.50  & 74.05 \\
 & Llama 3B       & 27.52 & 9.50  & 75.71 \\
 & Llama 8B       & 27.90 & 7.50  & 75.24 \\
 & +PL pretrain.& 42.42 & 10.50 & 74.76 \\
\midrule

\multirow{5}{*}{\rotatebox[origin=c]{90}{\textbf{En--Japanese}}}
 & Qwen 8B        & 7.77  & 3.50  & 43.81 \\
 & Llama 1B       & 23.15 & 6.00  & 0.00  \\
 & Llama 3B       & 12.57 & 6.50  & 25.71 \\
 & Llama 8B       & 14.92 & 4.50  & 0.24  \\
 & +JA pretrain.& 18.82 & 4.00  & 36.67 \\
\midrule

\multirow{4}{*}{\rotatebox[origin=c]{90}{\textbf{En--French}}}
 & Qwen 8B        & 26.60 & 8.00  & 87.38 \\
 & Llama 1B       & 36.05 & 5.50  & 89.05 \\
 & Llama 3B       & 36.03 & 6.50  & 90.48 \\
 & Llama 8B       & 46.75 & 7.00  & 89.29 \\
\midrule

\multirow{4}{*}{\rotatebox[origin=c]{90}{\textbf{En--Italian}}}
 & Qwen 8B        & 27.90 & 7.50  & 85.71 \\
 & Llama 1B       & 34.10 & 9.00  & 88.09 \\
 & Llama 3B       & 38.45 & 8.50  & 89.52 \\
 & Llama 8B       & 46.05 & 11.00 & 86.19 \\
\bottomrule
\end{tabular}
\caption{Conductivity measured using Accuracy (ACC) of fictional facts retrieval across language settings on the \bname and \clika fictional splits. We also report translation accuracy (exact match) for \clika.
}
\label{tab:fictional_combined}
\end{table}
    
\section{Experiments}
\label{sec:experiments}
\subsection{Experimental setup}

\paragraph{Models}
Multiple LLMs have been included in the experiments:\footnote{Due to hardware constraints, we were able to experiment with models of maximum 8B parameters.} 
three size variants of Llama 3 \cite[1B, 3B, 8B;][]{llama3.2,grattafiori2024llama3herdmodels}, Qwen 3 8B \cite{qwen3technicalreport} and the Polish LLM Bielik 4.5B v3 \cite{bielik-model-card, bielik-technical}. 
To inspect the influence of continued pretraining, we also included  Llama 8B versions further pretrained on Polish \cite{kocon2025pllum}, Japanese \cite{Fujii:COLM2024} and Chinese \cite{chinese-llama-alpaca}, denoted as Llama 8B +PL/JA/ZH pretrain. See App.~\ref{app:exsetup} for more details on the experimental setup.


\paragraph{Benchmarks}
The models were evaluated both by our \bname benchmark and by \clika~\cite{gao-etal-2024-multilingual} for comparison. 
Since \clika uses closed-form questions, we adapted it to open-form questions for better comparability.
Fine-tuning of models on fictional data was performed by AdamW optimizer with the default parameters. 
More details and basic dataset statistics are in App.~\ref{app:data-stat}. 


\paragraph{Evaluating the impact of cross-lingual alignment}
To examine how a state-of-the-art cross-lingual alignment method AFP~\cite{contrastive_learning_generative}  
influences knowledge transfer, experiments were conducted across three language settings: Llama 1B and 3B were aligned on English-Chinese and English-Polish pairs using translation and instruction data from OPUS100~\cite{opus} and BactrianX~\cite{bactrianx}, while an additional experiment used the Bielik Polish LLM with Polish as the source language aligned to Czech -- departing from the literature's typical English-centric pairs. The alignment process and its experimental verification is further described in App.~\ref{app:alignment-impact}.

\subsection{Results}
\paragraph{\bname is more challenging and provides more stable evaluation.}
Cross-lingual knowledge consistency results in Tab.~\ref{tab:factual_combined} show that \bname is more challenging than \clika; e.g., for English most LLMs exceed 80\% accuracy on \clika despite small model size while scores on \bname typically stay below 50\%. 
Accuracy across languages is also more stable on \bname, whereas \clika shows larger fluctuations. Knowledge consistency with English is high across all models on both benchmarks. Notably, continual pretraining on single-language data does not improve -- and in fact deteriorates -- consistency with English for all three languages tested.

\paragraph{Knowledge conductivity is insufficient in multi-lingual LLMs.}
Tab.~\ref{tab:fictional_combined} presents knowledge conductivity results. For \clika, we additionally report entity-name translation accuracy: while it is high for French and Italian, it is only moderate for Polish and low for Chinese and Japanese, raising validity concerns about \clika's conductivity scores. Conductivity as measured by \bname is consistently higher than \clika values (as \bname is not influenced by translation issues), yet the best result remains below 50\%, indicating insufficient conductivity overall. Romance languages (Italian, French) yield the highest scores across models. 
Notably, Qwen 8B -- reported to perform well on Chinese -- 
achieves a surprisingly low conductivity score on that language. In the App.~\ref{app:dates_cities}, we provide a more detailed table with conductivity scores reported for each type of question (dates, cities) separately.  To facilitate comparisons, we also provide a version of Tab.~\ref{tab:fictional_combined} with confidence intervals and standard errors in the App.~\ref{app:CI-TABS}. 

\paragraph{Continued pretraining improves conductivity} on two out of the three languages tested, as measured by \bname: gains exceed 50\% on Polish and 25\% on Japanese, while deterioration in Chinese is negligible. 
Results on \clika are harder to interpret, as some models without continual pretraining scored close to zero on translation accuracy.


\paragraph{Cross-lingual alignment substantially benefits knowledge conductivity} in  larger models (see Tab.~\ref{tab:afp_fic} in App.~\ref{app:alignment-impact}): Llama 3B and Bielik 4.5B more than double their Czech conductivity score and $>$70\% gains are observed on Polish and Chinese. Smaller gains are visible on \clika, likely due to low translation accuracy. The small Llama 1B model shows no improvement under AFP.
Note that the Polish-Czech pair was evaluated on \bname only, as \clika lacks Czech data. 

It is also important to note that AFP does not consistently improve performance on downstream evaluation tasks (see Tabs. \ref{tab:xnli_paws_llama_results_pl} and \ref{tab:xnli_paws_llama_results_zh} in App. \ref{app:alignment-impact}).
The evaluation on cross-lingual benchmarks, namely XNLI \cite{conneau-etal-2018-xnli} and PAWS-X \cite{yang-etal-2019-paws}, shows a downgrade after AFP on the smaller Llama 3.2 1B and more mixed results on the bigger 3B variant.

\section{Conclusions}
 This paper addresses the challenge of evaluating cross-lingual knowledge transfer in large language models by introducing \bname, a multilingual and easily extensible benchmark. The benchmark consists of two components: a dataset of factual information about people, which is used to assess knowledge consistency, and a dataset of fictional information about people, which, once injected into LLMs, is used to measure knowledge conductivity.

 The evaluation of cross-lingual knowledge consistency and conductivity revealed several trends. The proposed \bname benchmark is more challenging and yields more stable cross-lingual accuracies than the previous most prominent CLiKA benchmark. All models under study maintain high knowledge consistency with English. Continual monolingual pretraining generally decreases knowledge consistency but can significantly improve knowledge conductivity for some languages. 
 Cross-lingual alignment benefits larger models, as applying alignment techniques significantly boosts knowledge conductivity, e.g.\ doubling the scores for Czech.
  In contrast, for smaller models, cross-lingual alignment seems to not improve knowledge conductivity, but also does not enhance alignment accuracy in general.

\section*{Limitations}
Our benchmark focuses exclusively on knowledge transfer that can be expressed by RDF triples, more precisely on person-centric factual attributes (e.g., birth, death, location). These results may not extend to other types of knowledge, such as relational, procedural, or commonsense information.

Additionally, our benchmark relies on knowledge injection through model fine-tuning with QA pairs. Although this method has been used successfully in related research~\cite{gao-etal-2024-multilingual}, using a different knowledge injection procedure could yield different results.

Finally, our experiments are limited to models up to 8B parameters. As model scale can influence memorization and factual recall, our findings may not generalize to substantially larger models.

\section*{Ethical Considerations}
We acknowledge the use of GitHub Copilot in drafting parts of our code. Some portions of the manuscript were revised using an AI-assisted grammar checker.

\section*{Acknowledgments}
This research was partially funded by the European Union (ERC, NG-NLG, 101039303) and by the Charles University Research Center program No.~24/SSH/009.
This research was conducted using the computational infrastructure of the Poznan Supercomputing and Networking Center (PSNC) and resources of the LINDAT/CLARIAH-CZ Research Infrastructure (Czech Ministry of Education, Youth, and Sports project No. LM2023062).

\bibliography{compact}

\appendix
\section{Related Work}
\label{sec:related}


\paragraph{Large Language Models (LLMs)} have rapidly advanced following the 
introduction of the Transformer architecture \cite{GOAT-TRANSFORMER}, with 
decoder-only models now dominating the landscape \cite{Mistral-7B, grattafiori2024llama3herdmodels, 
qwen3technicalreport}. 
While early LLMs were focused on English~\cite{Llama}, the development of multilingual LLMs gained traction via mixed multilingual pretraining~\cite{Conneau}, i.e. including texts from other languages in the training corpora.
Notable examples include BLOOM~\cite{bloom}, Aya~\cite{ustun-etal-2024-aya} and mGPT~\cite{shliazhko-etal-2024-mgpt} which demonstrated that multilingual pretraining could yield competitive performance across diverse languages and narrows the gap with the monolingual counterparts.
More advanced forms of multilingual pretraining, in particular knowledge-aware pretraining to facilitate knowledge transfer~\cite{Jiang_Liang_Chen_Duan_2022}, have also been explored.
Despite these advances, lower-resourced languages continue to lag behind, as the training data imbalance between high- and low-resource languages remains a persistent challenge~\cite{ahuja-etal-2023-mega}.

\paragraph{Knowledge probing}
Factual knowledge in language models has long been studied with cloze-style, subject-relation-object triples. LAMA \cite{petroni-etal-2019-language} evaluates factual knowledge in pretrained language models, including BERT, by asking them to predict masked entities in English sentences. X-FACTR \cite{jiang-etal-2020-x} extends this to 23 languages and multi-token answers, while retaining a focus on masked language models. These benchmarks measure what a model knows when queried in a given language, not whether knowledge acquired in one language can be retrieved in another. 

\paragraph{Cross-lingual alignment} refers to meaningful similarity between text representations in different languages \cite{hammerl-etal-2024-understanding}. \citet{weak_strong_alignment_definitions} distinguish between \emph{weak alignment}, when nearest neighbors across languages correspond to translations; and \emph{strong alignment}, when same-meaning items are closer than unrelated ones in any language. Alignment can also be viewed functionally: a prediction head trained in one language should generalize across others given that the text representation provided by the model is well-aligned across languages~\cite{pires-etal-2019-multilingual}.

Alignment strategies can be roughly  divided into indirect and direct approaches, which are often combined in practice. 
\emph{Indirect alignment} leverages data augmentation and multilingual instruction finetuning to encourage shared representations without explicitly enforcing alignment. 
Techniques such as unsupervised machine translation~\cite{parallel-mining}, code-switching~\cite{code_switching_pretraining}, and specialized data augmentations~\cite{weam} mix bilingual data with modified objectives to indirectly improve multilingual consistency. 
While many of these methods are designed for encoder-only models, Cross-lingual Instruction Following (CIF) aligns decoder-only models by finetuning it with multilingual instruction–response pairs. 

\emph{Direct alignment} explicitly optimizes for representational similarity, typically via contrastive learning, which pulls together parallel or semantically equivalent pairs while pushing apart unrelated samples. 
Approaches such as PreAlign~\cite{contrastive_prealign} and mRASP2~\cite{contrastive_machine_translation} integrate contrastive objectives with code-switched or augmented data to strengthen multilingual consistency. 
The Align after Pretraining (AFP, \citealp{contrastive_learning_generative}) method exemplifies the synergy of both strategies, combining CIF-style instruction finetuning with contrastive learning to align sentence-level representations in decoder-only models. 
AFP is used in our experiments as a state-of-the-art cross-lingual alignment method.  

\paragraph{Alignment evaluation}
Evaluation of cross-lingual alignment can be performed by directly quantifying the similarity between representations in different languages or by measuring the model performance on cross-lingual transfer tasks. 
These include retrieval-based benchmarks like LAReQA~\cite{weak_strong_alignment_definitions} that test whether models can identify relevant information across languages and zero-shot evaluation where the model is fine-tuned on a given task (e.g. natural language inference) with a source language data and then tested on target language data~\cite{contrastive_learning_generative,contrastive_prealign}.
Recent approaches~\cite{ifergan-etal-2025-beneath} use model editing methods to verify whether changes propagates to other languages through shared representations.

Since it is difficult to determine which knowledge the model acquired in English and/or in the target languages, the evaluation of cross-lingual knowledge conductivity is difficult and requires injecting synthetic data that the model is known to have encountered only in a single language~\cite{gao-etal-2024-multilingual}.
One of the approaches that evaluates the knowledge transfer is CLKA (Cross-lingual Knowledge Application, \citealp{contrastive_prealign}), which trains the model on English data describing synthetic knowledge expressed in RDF triples, and then tests the knowledge retention in a target language. This method was further developed into \clika~\cite{gao-etal-2024-multilingual}, where the knowledge transfer of mixed pretraining and continual pretraining was tested via question answering.

\section{Problems of \clika}
\label{sec:clika}

To measure knowledge conductivity, \clika performs model fine-tuning on QA pairs related to synthetic facts in a source language. The model is also fine-tuned to translate fact-related entity names in both source and target languages. Following this finetuning, the model is tested on its ability to answer question about synthetic facts in the source language. 

We believe that the inclusion of the translation task in the \clika methodology could lead to introducing additional noise into the results. 
First, we found that the translation of place names in the CLiKA dataset was particularly error-prone. 
We found that different English entity names were sometimes translated into a single entity name in the target language, creating a mismatch between information conveyed in both languages. Some English names of fictional places were occasionally translated into names of real places, which violates the assumption that the model has not seen the entities during pretraining.
Next, we argue that the method not only measures the models ability of retrieving knowledge learn in a different language, but also the ability to correctly translate proper names, which may influence the results. 
This also creates challenges in extending \clika to new languages, as one needs to prepare the translations of all sentences and entity names.
In practice, we found that for some model-language pairs, fine-tuning failed to produce translation quality sufficient for meaningful evaluation (see Sec.~\ref{sec:experiments}).

\section{Dataset construction}
\label{app:data-construction}
\paragraph{Factual data}
The database of factual people was obtained by sampling Wikidata \cite{wikidata}.
Since our database includes names, birth dates, death dates and a places of birth, we filtered the entities to only those containing this information. 
More specifically, we filtered for people whose entries include the “birthdate”, “deathdate”, and “multilingual name” data fields, and whose “place of birth” is of the entity type “city”.
We randomly selected 400 people passing our filters, and we added to the database all associated RDF triples containing the selected relations/fields.

\paragraph{Fictional data}
For the construction of the fictional database of \bname, we opted for Polish names and cities, as the resources were easily available and they are often not translated into other languages and used as-is.
To create life-like personas, the lists of names and surnames provided by the Polish government were used \cite{ministerstwo2023imiona,ministerstwo2025nazwiska}.  The cities (places of residence) were collected from the SimpleMaps database \cite{simplemaps2025plcities}. 

To avoid non-standard  or foreign names/\hspace{0mm}surnames, we excluded low-frequency entries from the lists. The threshold for filtering names and surnames was computed as:
\[
T_{\text{name}} = 0.1N_{\max\_n} 
\]
\[
 T_{\text{surname}} = 0.05N_{\max\_s} 
\]
where $N_{max\_n}$ ($N_{max\_s}$) represents the number of occurrences of the most frequent name (surname).
Then, from the filtered lists, 400 names and surnames were sampled independently to initiate creation of entities. Each entity formed three RDF triples with these relations: birth date, death date, and place of living. The years of birth were generated using a uniform distribution, with the boundaries of 1800 and 1940. Life expectancy was sampled from  normal distribution with the $\mu=65$ and $\sigma = 10$. The places of living were randomly assigned to each person from the list of the 20 most populated cities in Poland.

\paragraph{Templates construction}
To convert the constructed databases of individuals into QA pairs, we used text templates (see Sec.~\ref{sec:methodology}). 
For each type of relation (i.e., birth, death, place of birth), we constructed 10 QA templates for fictional entities using GPT-4 for each of the considered languages (see prompts in App.~\ref{apendix-a}). For the factual part, one template for each relation was crafted by hand and translated to other languages, since it required a more precise formulation.
The templates were validated by humans and edited if needed. In total, 30 templates per language were constructed.

\paragraph{Template generation prompts}
\label{apendix-a}
The prompts used to automatically generate question-answer templates are presented in Tab.~\ref{tab:t1}, \ref{tab:t2} and \ref{tab:t3}.
As an example, the generated templates used to generate questions and answers for fictional part of English dataset are presented in Tabs.~\ref{tab:a1} and~\ref{tab:a2}, respectively. The templates for factual part are presented in Tab.~\ref{tab:a3}.

\section{Experimental setup}
\label{app:exsetup}
\paragraph{Models}
To comprehensively assess the cross-lingual knowledge conductivity in current LLMs, multiple modern models have been included in the experiments:\footnote{Due to hardware constraints, we were able to experiment with max.~8B models.} 
two size variants (1B and 3B) of Llama 3.2 \cite{llama3.2}, larger Llama 3 8B \cite{grattafiori2024llama3herdmodels}, Qwen 3 8B \cite{qwen3technicalreport} and the Polish LLM Bielik 4.5B v3 \cite{bielik-model-card, bielik-technical}. 
Moreover, to inspect the influence of continued pretraining on the knowledge conductivity, we included Llama 8B models additionally pretrained on Polish (Pllum Llama 8B based on Llama 3.1, \citealp{kocon2025pllum}), Japanese (Llama-3-Swallow 8B, \citealp{Fujii:COLM2024}), and Chinese (Llama-3-Chinese 8B, \citealp{chinese-llama-alpaca}). 
These models are later denoted as Llama 8B +PL pretain., +JA pretain. and +ZH pretain., respectively.

\paragraph{Benchmarks}
The models were evaluated both by our \bname benchmark and by CLiKA~\cite{gao-etal-2024-multilingual} for comparison.
Since CLiKA evaluates factual knowledge in a closed question manner, the method has been adapted to answer questions by text completion to be fully comparable with the factual part of \bname. The modification was straightforward, as the questions was answerable even if omited the list of possible answers.
To address the ambiguity of the generated answers evaluation method, a liberal approach was used that identified whether the answer included the tag, continent, or translation in the case of CLiKA, and a date or city in the case of \bname. The answer does not have to include the subject of the question to be correct, as this method is prone to misclassifying the answer as incorrect when the retrieved entity name is conjugated. 

LLM answers were generated by greedy decoding to provide better reproducibility.
Fine-tuning of models on fictional data was performed by AdamW optimizer with the default parameters in the HuggingFace library~\cite{wolf-etal-2020-transformers}. The process of fine-tuning ends when the model reaches a 90\% accuracy threshold on the validation data.
To validate the experimental setup, we examined the possible relationship between the validation accuracy reached in the knowledge injection phase on the resulting conductivity scores.
We computed correlations scores between validation accuracy reached (i.e. the accuracy after crossing the 90\% threshold) with the resulting conductivity scores across the conducted experiments. The Pearson correlation is $-0.175$ for \bname and $0.203$ for \clika (see Tab. \ref{tab:correlation_val}), which does not provide evidence of a statistically significant relationship between the two. 

\begin{table}[ht]
\centering
\small
\label{tab:pearson_corr}
\begin{tabular}{lrr}
\hline
\textbf{Splits} & \textbf{Pearson $r$} & \textbf{$n$} \\
\hline
CLKTx Fic. ACC vs Val. CLKTx Fic. & $-0.175$ & 23 \\
CLIKA Fic. ACC vs Val. CLIKA Fic. & $0.203$ & 23 \\
\hline
\end{tabular}
\caption{Pearson correlation between conductivity evaluation on \bname and \clika, and the validation scores when the knowledge injection phase reached its accuracy threshold.}
\label{tab:correlation_val}
\end{table}

\section{Dataset statistics}
\label{app:data-stat}
Basic statistics of both fictional datasets used for knowledge injection (fine-tuning) used in \bname are presented in Tabs.~\ref{tab:clika-stats} and~\ref{tab:fic-stats}. 
\begin{table}[t]
\small
\centering
\begin{tabular}{lcc}
\toprule
\textbf{CLiKA Fictional} & \textbf{Src Lang} & \textbf{Tgt Lang} \\
\midrule
Test Facts              & 0     & 200   \\
Test Translations *     & 0     & 420   \\
Validation Facts        & 400   & 200   \\
Validation Translations * & 4200  & 0     \\
Training Facts          & 4000  & 2000  \\
Training Translations * & 4200  & 0     \\
\bottomrule
\end{tabular}

\caption{Distribution of samples across source and target languages in CLiKA knowledge injection. *Since the translation includes two languages, samples are categorized according to the translation direction, that is, by the resulting language of the translation.}
\label{tab:clika-stats}
\end{table}

\begin{table}[t]
\small
\centering
\begin{tabular}{lcc}
\toprule
\textbf{\bname Fictional} & \textbf{Src Lang} & \textbf{Tgt Lang} \\
\midrule
Test        & 0      & 6000  \\
Validation  & 600    & 600   \\
Training    & 12000  & 6000  \\
\bottomrule
\end{tabular}

\caption{Distribution of samples across source and target languages during \bname knowledge injection.}
\label{tab:fic-stats}
\end{table}

\section{Impact of cross-lingual alignment on knowledge conductivity}
\label{app:alignment-impact}

\subsection{Data preparation and alignment process}
To examine whether a state-of-the-art cross-lingual alignment method, AFP~\cite{contrastive_learning_generative}, influences the cross-lingual knowledge transfer, an experiment concerning three language settings was conducted. 
Llama 3.2 1B and 3B models were aligned on English-Chinese and English-Polish language pairs, using 
translation and instruction data from the OPUS100    \cite{opus} and BactrianX \cite{bactrianx} datasets. 

Since most cross-lingual alignment experiments presented in the literature concern pairs of languages involving English, we also conducted an experiment where Polish is treated as a source language in the Bielik Polish LLM~\cite{bielik-model-card}, and an alignment to the Czech language was conducted. 
Cross-lingual alignment was performed as previously by the AFP method, but the WikiMatrix parallel corpus \cite{wikimatrix} for the Polish–Czech pair was used instead of OPUS-100, as the latter is centered around English.

The batch parameters for AFP were chosen accordingly to the size of the used model, we have maximized the per-device batch size and batch size. The total batch size was kept equal to 128 by modifying the gradient accumulation steps. The alignment procedure in each case lasted 10,000 steps.
The AdamW optimizer parameters were set to $\beta_1=0.9$ and $\beta_1=0.999$, and the learning rate to $1e-5$. The $\beta$ weight associated with the CLM loss was set to 1.5,  while the $\alpha$ CL weight was set to 1, as those were default parameters provided in the authors' repository. The temperature parameter present in contrastive learning was set to  $\tau=0.05$. The layer chosen for the mean pooling of the sentence embeddings was set to 1, following the setup of the original experiments \cite{contrastive_learning_generative}. 
The experiments were performed on two Nvidia A100 80 GB GPUs.

To verify the extent of cross-lingual alignment achieved by AFP in our experiments, we measured the performance of two multilingual reasoning tasks:  natural language inference  on XNLI dataset~\cite{conneau-etal-2018-xnli} and paraphrase detection on PAWS-X dataset \cite{yang-etal-2019-paws}. 

As XNLI and PAWS-X do not include Czech, a simple translation pipeline was designed: the original datasets were translated into Czech by an MT system and a larger LLM was applied to check the labeling consistency with the original datasets. We used  EuroLLM 9B \cite{euro-llm} for data translation and Qwen 2.5 72B \cite{qwen2} for consistency assurance (see prompts in App.~\ref{apendix-b}).
Note that XNLI and PAWS-X are evaluated on English using two models aligned with different language pairs (English–Chinese and English–Polish). Therefore, the results naturally differ and do not indicate  inconsistencies in our experimental setup.

\subsection{Alignment verification}

Before discussing the impact of cross-lingual alignment on knowledge conductivity, we first present the results of alignment evaluation between English and Chinese (Tab. \ref{tab:xnli_paws_llama_results_zh}), English and Polish (Tab. \ref{tab:xnli_paws_llama_results_zh}), and Polish and Czech (Tab. \ref{tab:xnli_paws_bielik_results}).
The results do not present a clear trend of improvements after the application of the AFP cross-lingual alignment method -- the accuracy changes are inconsistent across models and languages.
Nevertheless, for the larger Llama model, AFP improved 5-shot performance across both tasks and all target languages.
In this setup also the 5-shot performance on the source language (English) was improved with the only exception of PAWS-X and Polish. 

\subsection{Knowledge conductivity evaluation}
The results of knowledge conductivity evaluation are presented in Tab.~\ref{tab:afp_fic} on both CLiKA and \bname benchmarks. As CLiKA is lacking Czech data, the Polish-Czech pair was only evaluated by the proposed \bname.
For the larger models, Llama 3B and Bielik 4.5B, AFP significantly improved knowledge conductivity, more than doubling the result on Czech and producing $>70\%$ increases on Polish and Chinese. 
The improvement is also visible on the CLiKA benchmark, but the reported values are lower, probably due to low translation accuracy scores.
In contrast, AFP applied to the smaller 1B Llama model did not yield any improvements.
\begin{table}[t]
\small
\centering
\setlength{\tabcolsep}{3pt} 
\begin{tabular}{lc|cc}
\toprule
\textbf{Model} & \textbf{\bname} & \multicolumn{2}{c}{\textbf{CLiKA}} \\
 & \textbf{Fic. ACC}  & \textbf{Fic. ACC} & \textbf{Transl. ACC} \\
\midrule
\multicolumn{4}{c}{\textbf{English--Chinese}} \\
\midrule
Llama 1B & 24.90 & 5.50 & 0.00 \\
\quad + AFP & 19.90 & 5.50 & 10.47 \\
\midrule
Llama 3B & 18.32 & 3.50 & 1.67 \\
\quad + AFP & 31.42 & 6.00 & 21.19 \\
\midrule
\multicolumn{4}{c}{\textbf{English--Polish}} \\
\midrule
Llama 1B & 37.08 & 5.50 & 74.05 \\
\quad + AFP & 35.22 & 3.50 & 65.93 \\
\midrule
Llama 3B & 27.52 & 9.50 & 75.71 \\
\quad + AFP & 47.67 & 10.50 & 70.23 \\
\midrule
\multicolumn{4}{c}{\textbf{Polish--Czech}} \\
\midrule
Bielik 4.5B & 15.40 & n/a & n/a \\
\quad + AFP & 37.45 & n/a & n/a \\
\bottomrule
\end{tabular}
\caption{Impact of AFP alignment on knowledge conductivity across different language settings, measured by fictional facts retrieval accuracy, using Llama 3.2 (1B, 3B) and Bielik 4.5 B v3 models. 
The CLiKA column additionally reports accuracy of fictional names translation. Missing values in the Bielik setting are due to the absence of a Czech split in the CLiKA dataset.}
\label{tab:afp_fic}
\end{table}

\section{Detailed results}
\paragraph{The results of the fictional part of \bname across language-specific and language-agnostic answer types.}
The evaluation results split across types of answers: those concerning Polish cities (place of living) and language-agnostic dates (year of birth and year of death) are presented in Tab.~\ref{tab:clktx-fic-acc-splits}. While this distinction changes the ordering of the scores, the highest accuracy on the cities split is not achieved in Polish. Similarly, the high conductivity scores for the English-Polish pair are maintained after the split, so they are not result of using Polish for data generation. Hence, the bias introduced by the use of Polish city names seems limited.
\label{app:dates_cities}

\paragraph{\bname and \clika with standard error and confidence intervals}
\label{app:CI-TABS}
The results reported together with standard error and confidence intervals in Tab. \ref{tab:ordered-ci-table} show the resolution advantage of \bname over \clika. Given the size of confidence intervals, results reported using \clika are almost undistinguishable, whereas \bname proves to produce less noisy, more stable evaluation.

\section{Translation and verification prompts of cross-lingual evaluation benchmarks}
\label{apendix-b}
The prompts used to translate XNLI and PAWS-X are presented in Tab.~\ref{tab:a4}. The prompts used to verify the correctness of translated content are presented in Tab.~\ref{tab:a5}.
\begin{table*}[t]
\centering
\small
\setlength{\tabcolsep}{6pt}
\begin{tabular}{llccc}
\toprule
Language Pair & Model & \bname Fic. Acc & \bname Fic. Acc dates & \bname Fic. Acc cities \\
\midrule
\multirow{5}{*}{En-Chinese} & Qwen 8B & 16.15 & 10.38 & 27.70 \\
 & Llama 1B & 24.90 & 17.72 & 39.25 \\
 & Llama 3B & 18.32 & 9.90 & 35.15 \\
 & Llama 8B & 26.55 & 18.38 & 42.90 \\
 & +ZH pretrain & 25.95 & 13.73 & 50.40 \\
\midrule
\multirow{5}{*}{En-Polish} & Qwen 8B & 22.85 & 14.20 & 40.15 \\
 & Llama 1B & 37.08 & 31.27 & 48.70 \\
 & Llama 3B & 27.52 & 16.48 & 49.60 \\
 & Llama 8B & 27.90 & 18.27 & 47.15 \\
 & +PL pretrain & 42.43 & 32.38 & 62.55 \\
\midrule
\multirow{5}{*}{En-Japanese} & Qwen 8B & 7.77 & 5.15 & 13.00 \\
 & Llama 1B & 23.15 & 15.47 & 38.50 \\
 & Llama 3B & 12.57 & 3.60 & 30.50 \\
 & Llama 8B & 14.92 & 8.75 & 27.25 \\
 & +JA pretrain & 18.82 & 4.98 & 46.50 \\
\midrule
\multirow{4}{*}{En-French} & Qwen 8B & 26.60 & 18.88 & 42.05 \\
 & Llama 1B & 36.05 & 32.95 & 42.25 \\
 & Llama 3B & 36.03 & 26.70 & 54.70 \\
 & Llama 8B & 46.77 & 37.10 & 66.10 \\
\midrule
\multirow{4}{*}{En-Italian} & Qwen 8B & 27.80 & 19.50 & 44.40 \\
 & Llama 1B & 34.10 & 30.12 & 42.05 \\
 & Llama 3B & 38.45 & 26.45 & 62.45 \\
 & Llama 8B & 46.05 & 40.25 & 57.65 \\
\bottomrule
\end{tabular}
\caption{\bname fictional split accuracy with numeric (dates) and non-numeric (cities) splits.}
\label{tab:clktx-fic-acc-splits}
\end{table*}
\begin{table*}[t]
\centering
\small
\setlength{\tabcolsep}{4pt}
\begin{tabular}{llcclccl}
\toprule
 & & \multicolumn{3}{c}{\bname Fic.} & \multicolumn{3}{c}{CLIKA Fic.} \\
\cmidrule(lr){3-5} \cmidrule(lr){6-8}
Language Pair & Model & ACC & SE & 95\% CI & ACC & SE & 95\% CI \\
\midrule
\multirow{5}{*}{En-Chinese} & Qwen 8B & 16.15 & 0.48 & [15.22, 17.08] & 5.50 & 1.61 & [2.34, 8.66] \\
 & Llama 1B & 24.90 & 0.56 & [23.81, 25.99] & 5.50 & 1.61 & [2.34, 8.66] \\
 & Llama 3B & 18.32 & 0.50 & [17.34, 19.30] & 3.50 & 1.30 & [0.95, 6.05] \\
 & Llama 8B & 26.55 & 0.57 & [25.43, 27.67] & 7.50 & 1.86 & [3.85, 11.15] \\
 & +ZH pretrain. & 25.95 & 0.57 & [24.84, 27.06] & 5.50 & 1.61 & [2.34, 8.66] \\
\midrule
\multirow{5}{*}{En-Polish} & Qwen 8B & 22.85 & 0.54 & [21.79, 23.91] & 6.00 & 1.68 & [2.71, 9.29] \\
 & Llama 1B & 37.08 & 0.62 & [35.86, 38.30] & 5.50 & 1.61 & [2.34, 8.66] \\
 & Llama 3B & 27.52 & 0.58 & [26.39, 28.65] & 9.50 & 2.07 & [5.44, 13.56] \\
 & Llama 8B & 27.90 & 0.58 & [26.77, 29.03] & 7.50 & 1.86 & [3.85, 11.15] \\
 & +PL pretrain. & 42.42 & 0.64 & [41.17, 43.67] & 10.50 & 2.17 & [6.25, 14.75] \\
\midrule
\multirow{5}{*}{En-Japanese} & Qwen 8B & 7.77 & 0.35 & [7.09, 8.45] & 3.50 & 1.30 & [0.95, 6.05] \\
 & Llama 1B & 23.15 & 0.54 & [22.08, 24.22] & 6.00 & 1.68 & [2.71, 9.29] \\
 & Llama 3B & 12.57 & 0.43 & [11.73, 13.41] & 6.50 & 1.74 & [3.08, 9.92] \\
 & Llama 8B & 14.92 & 0.46 & [14.02, 15.82] & 4.50 & 1.47 & [1.63, 7.37] \\
 & +JA pretrain. & 18.82 & 0.50 & [17.83, 19.81] & 4.00 & 1.39 & [1.28, 6.72] \\
\midrule
\multirow{4}{*}{En-French} & Qwen 8B & 26.60 & 0.57 & [25.48, 27.72] & 8.00 & 1.92 & [4.24, 11.76] \\
 & Llama 1B & 36.05 & 0.62 & [34.84, 37.26] & 5.50 & 1.61 & [2.34, 8.66] \\
 & Llama 3B & 36.03 & 0.62 & [34.82, 37.24] & 6.50 & 1.74 & [3.08, 9.92] \\
 & Llama 8B & 46.75 & 0.64 & [45.49, 48.01] & 7.00 & 1.80 & [3.46, 10.54] \\
\midrule
\multirow{4}{*}{En-Italian} & Qwen 8B & 27.90 & 0.58 & [26.77, 29.03] & 7.50 & 1.86 & [3.85, 11.15] \\
 & Llama 1B & 34.10 & 0.61 & [32.90, 35.30] & 9.00 & 2.02 & [5.03, 12.97] \\
 & Llama 3B & 38.45 & 0.63 & [37.22, 39.68] & 8.50 & 1.97 & [4.63, 12.37] \\
 & Llama 8B & 46.05 & 0.64 & [44.79, 47.31] & 11.00 & 2.21 & [6.66, 15.34] \\
\bottomrule
\end{tabular}
\caption{Performance table reporting accuracy, standard errors, and 95\% confidence intervals for fictional splits of \bname  and \clika }
\label{tab:ordered-ci-table}
\end{table*}
\begin{table*}[t]
\small
\centering
\begin{tabular}{l>{\hspace{3mm}}cccc>{\hspace{3mm}}cccc}
\toprule
\textbf{Model} & \multicolumn{4}{c}{\textbf{XNLI}} & \multicolumn{4}{c}{\textbf{PAWS-X}} \\
 & En@0 & En@5 & Zh@0 & Zh@5 & En@0 & En@5 & Zh@0 & Zh@5 \\
\midrule
Llama 3.2 1B & 54.29 & 50.82 & 45.23 & 39.48 & 56.20 & 54.00 & 54.80 & 54.90 \\
Llama 3.2 1B + AFP & 48.70 & 46.45 & 39.22 & 38.86 & 57.90 & 59.50 & 55.35 & 54.35 \\
\midrule
Llama 3.2 3B & 53.67 & 43.03 & 50.50 & 36.47 & 69.70 & 64.30 & 59.60 & 54.85 \\
Llama 3.2 3B + AFP & 54.21 & 48.84 & 43.85 & 45.59 & 64.10 & 64.30 & 57.85 & 57.55 \\
\bottomrule
\end{tabular}
\caption{Results of Llama 3.2 1B and 3B models with and without AFP on XNLI and PAWS-X datasets for English (En) and Chinese (Zh) at 0-shot and 5-shot settings.}
\label{tab:xnli_paws_llama_results_zh}
\end{table*}

\begin{table*}[t]
\small
\centering
\begin{tabular}{l>{\hspace{3mm}}cccc>{\hspace{3mm}}cccc}
\toprule
\textbf{Model} & \multicolumn{4}{c}{\textbf{XNLI}} & \multicolumn{4}{c}{\textbf{PAWS-X}} \\
 & En@0 & En@5 & Pl@0 & Pl@5 & En@0 & En@5 & Pl@0 & Pl@5 \\
\midrule
Llama 3.2 1B & 54.29 & 50.82 & 43.08 & 44.03 & 56.20 & 54.00 & 61.83 & 45.93 \\
Llama 3.2 1B + AFP & 49.22 & 46.77 & 42.82 & 43.62 & 53.05 & 58.40 & 54.07 & 51.04 \\
\midrule
Llama 3.2 3B & 53.67 & 43.03 & 43.41 & 37.61 & 69.70 & 64.30 & 71.61 & 46.94 \\
Llama 3.2 3B + AFP & 52.22 & 49.90 & 41.36 & 47.68 & 66.45 & 61.95 & 57.67 & 49.84 \\
\bottomrule
\end{tabular}
\caption{Results of Llama 3.2 1B and 3B models with and without AFP on XNLI and PAWS-X datasets for English (En) and Polish (Pl) at 0-shot and 5-shot settings.}
\label{tab:xnli_paws_llama_results_pl}
\end{table*}

\begin{table*}[t]
\small
\centering
\begin{tabular}{l>{\hspace{3mm}}cccc>{\hspace{3mm}}cccc}
\toprule
\textbf{Model} & \multicolumn{4}{c}{\textbf{XNLI}} & \multicolumn{4}{c}{\textbf{PAWS-X}} \\
 & Pl@0 & Pl@5 & Cs@0 & Cs@5 & Pl@0 & Pl@5 & Cs@0 & Cs@5 \\
\midrule
Bielik 4.5B & 52.86 & 57.18 & 44.77 & 45.77 & 79.87 & 82.59 & 73.16 & 78.93 \\
Bielik 4.5B + AFP & 51.80 & 43.60 & 46.51 & 43.79 & 82.96 & 77.72 & 71.00 & 75.89 \\
\bottomrule
\end{tabular}
\caption{Results of Bielik 4.5B model with and without alignment method on XNLI and PAWS-X datasets for Polish (Pl) and Czech (Cs) at 0-shot and 5-shot settings.}
\label{tab:xnli_paws_bielik_results}
\end{table*}

\begin{table*}
\centering
\small
\begin{tabular}{p{2cm}p{13cm}}
\toprule
\textbf{Prompt} & I am creating a fictional dataset, and I need 10 templates to be filled with information about birth data. The template must have a place to fill in the date \{date\} and name \{name\} of the person. The date consists only of the year. Create 10 different \textbf{Target language} prompt formats consisting of a question and an answer and return them in JSON. The questions should only ask about the year of birth. The answer should be the same in each template. Example of a filled template: \{ "Birth": [ \{ "question": "In what year was John Smith born?", "answer": "John Smith was born in 1980." \}, ] \\ 
\bottomrule
\end{tabular}
\caption{Questions and answers template generation about a person's year of birth}
\label{tab:t1}
\end{table*}

\begin{table*}
\centering
\small
\begin{tabular}{p{2cm}p{13cm}}
\toprule
\textbf{Prompt} & I am creating a fictional dataset, and I need 10 templates to be filled with information about death data. The template must have a place to fill in the date \{date\} and name \{name\} of the person. The date consists only of the year. Create 10 different \textbf{Target language} prompt formats consisting of a question and an answer and return them in JSON. The questions should only ask about the year of death. The answer should be the same in each template. Example of a filled template: \{ "Death": [ \{ "question": "In what year did John Smith die?", "answer": "John Smith died in 1980." \}, ] \\ 
\bottomrule
\end{tabular}
\caption{Questions and answers template generation about a person's year of death}
\label{tab:t2}
\end{table*}

\begin{table*}
\centering
\small
\begin{tabular}{p{2cm}p{13cm}}
\toprule
\textbf{Prompt} & I am creating a fictional dataset, and I need 10 templates to be filled with information about people and their place of living. The template must have a place to fill in the location \{location\} and name \{name\} of the person. The location consists only of a city name. Create 10 different \textbf{Target language} prompt formats consisting of a question and an answer and return them in JSON. The questions should ask only about the place of living during the lifetime. The answer should be the same in each template. Example of a filled template: \{ "Place of living": [ \{ "question": "What city did John Smith live in?", "answer": "John Smith lived in New York." \}, ] \\ 
\bottomrule
\end{tabular}
\caption{Questions and answers template generation about a person's place of living}
\label{tab:t3}
\end{table*}

\begin{table*}[t]
\centering
\small
\begin{tabular}{@{}>{\bfseries}p{2.2cm} p{12.5cm}@{}}
\toprule
\textbf{Question type} & \textbf{Question} \\
\midrule
\multirow{10}{*}{Birth}
& In what year was \{name\} born? \\
& Can you tell me the year \{name\} was born? \\
& What is the birth year of \{name\}? \\
& Do you know when \{name\} was born? \\
& Which year marks the birth of \{name\}? \\
& When did \{name\} come into the world? \\
& Could you specify the year \{name\} was born? \\
& Please provide the birth year of \{name\}. \\
& I'd like to know when \{name\} was born. What year was it? \\
& When exactly was \{name\} born (just the year)? \\
\midrule
\multirow{10}{*}{Death}
& In what year did \{name\} die? \\
& What year marks the death of \{name\}? \\
& Can you tell me the year \{name\} passed away? \\
& When did \{name\} die? \\
& Which year did \{name\} pass away? \\
& Do you know the year of \{name\}'s death? \\
& Could you tell me when \{name\} died? \\
& In which year did \{name\} die? \\
& What year did \{name\} die? \\
& Please provide the year when \{name\} passed away. \\
\midrule
\multirow{10}{*}{Place of living}
& What city did \{name\} live in? \\
& Can you tell me the city where \{name\} lived? \\
& Which city was home to \{name\}? \\
& Do you know the city where \{name\} used to live? \\
& In what city did \{name\} live? \\
& What was the name of the city where \{name\} lived? \\
& Where in the city did \{name\} live? \\
& Can you name the city where \{name\} lived? \\
& Tell me the city \{name\} lived in. \\
& Which city was \{name\}'s home? \\
\bottomrule
\end{tabular}
\caption{Generated question templates regarding the year of birth, year of death and place of living in English}
\label{tab:a1}
\end{table*}

\begin{table*}
\centering
\small
\begin{tabular}{ll}
\toprule
\textbf{Category} & \textbf{Answer Template} \\
\midrule
Birth           & \{name\} was born in \{date\}. \\
Death           & \{name\} died in \{date\}. \\
Place of living & \{name\} lived in \{location\}. \\
\bottomrule
\end{tabular}
\caption{Generated answer templates for English}
\label{tab:a2}
\end{table*}
\begin{table*}
\centering
\small
\begin{tabular}{p{3cm} p{5cm} p{5cm}}
\toprule
\textbf{Category} & \textbf{Question} & \textbf{Answer} \\
\midrule
Birth & In what year was \{name\} born? & \{name\} was born in \{date\}. \\
Death & In what year did \{name\} die? & \{name\} died in \{date\}. \\
Place of birth & In which city was \{name\} born? & The city of \{name\}'s birth is \{location\}. \\
\bottomrule
\end{tabular}
\caption{Factual question and answer templates for English}
\label{tab:a3}
\end{table*}

\begin{table*}
\centering
\small
\begin{tabular}{p{3cm}p{10cm}}
\toprule
\textbf{Prompt Type} & \textbf{Prompt} \\
\midrule
\textbf{XNLI} & You are a translation engine. Your only task is to translate pairs of sentences. Translate from \texttt{\{lang\_in\}} to \texttt{\{lang\_out\}}, using the \texttt{\{lang\_out\}} alphabet. Preserve the original meaning of both sentences. Your response must consist of only the two translated sentences in \texttt{\{lang\_out\}}, each starting with 'Sentence 1:' and 'Sentence 2:'. No extra output. \\
\midrule
\textbf{PAWS-X} & You are a translation engine. Your only task is to translate pairs of sentences. Translate from \texttt{\{lang\_in\}} to \texttt{\{lang\_out\}}, using the \texttt{\{lang\_out\}} alphabet. These sentences are\texttt{\{negation\}} paraphrases and should be translated into \texttt{\{lang\_out\}} as such. Your response must consist of only the two translated sentences in \texttt{\{lang\_out\}}, each starting with 'Sentence 1:' and 'Sentence 2:'. No extra output. \\
\bottomrule
\end{tabular}
\caption{Prompts used for translating XNLI and PAWS-X}
\label{tab:a4}
\end{table*}

\begin{table*}
\centering
\small
\begin{tabular}{p{3cm}p{10cm}}
\toprule
\textbf{Prompt Type} & \textbf{Prompt} \\
\midrule
\textbf{XNLI} & You are a natural language inference engine. Given two sentences in \texttt{\{lang\_out\}}, determine their logical relationship. If the first sentence entails the second, return entailment. If the first sentence contradicts the second, return contradiction. If they are unrelated, return neutral. Respond with exactly one of: entailment, contradiction, or neutral. No explanations. \\
\midrule
\textbf{PAWS-X } & You are a paraphrase detection engine. Given two sentences in \texttt{\{lang\_out\}}, determine whether they are paraphrases. If the sentences are paraphrases return True, if not return False. No explanations. \\
\bottomrule
\end{tabular}
\caption{Verification prompts for XNLI and PAWS-X tasks. The verification is performed by classification of translated examples and checking its label with the gold standard.}
\label{tab:a5}
\end{table*}
\end{document}